\documentclass[10pt,twocolumn,letterpaper]{article}
\usepackage[T1]{fontenc}
\usepackage{times}
\usepackage[letterpaper,hmargin=0.75in,top=0.73in,bottom=0.82in,footskip=0.37in]{geometry}
\usepackage[numbers,sort&compress]{natbib}
\usepackage[font=small,labelfont=bf,labelsep=period]{caption}

\usepackage{graphicx}
\ifdefined\pdfminorversion\pdfminorversion=7\fi
\usepackage{amsmath,amssymb}
\usepackage{booktabs}
\usepackage{multirow}
\usepackage{microtype}
\usepackage{enumitem}
\usepackage{xcolor}
\definecolor{linkblue}{rgb}{0.10,0.25,0.42}
\usepackage[breaklinks,colorlinks,hyperfootnotes=false,allcolors=linkblue]{hyperref}
\hypersetup{pdftitle={LATER: Learning Action-Conditioned Temporal Evidence Retrieval for Predictive World Models},
            pdfauthor={Zihan Zhou, Wenxi Xu, Qilin Wang, Hao Tang},
            pdflang={en}}
\input{glyphtounicode}
\title{\bfseries LATER: Learning Action-Conditioned Temporal Evidence Retrieval\\
for Predictive World Models}
\author{Zihan Zhou \qquad Wenxi Xu \qquad Qilin Wang \qquad Hao Tang\textsuperscript{*}}
\date{}

\begin{document}
\maketitle
\begingroup
\renewcommand{\thefootnote}{*}
\footnotetext{Corresponding author.}
\endgroup

\begin{abstract}
Early interactions can reveal physical properties that later observations no longer show. These properties can still govern the effects of future actions. Inspired by the distinction between retention and selective readout in working memory and cognitive science, we introduce \textbf{LATER}, an action-conditioned world model built on CF-JEPA. LATER stores completed real transitions in partitioned memories that remain fixed during candidate rollouts. A response bridge queries these memories using the current state and candidate action, while multi-cut auxiliary supervision trains the memory writers. On distracted Two-Stage PushT, LATER raises terminal success from $15.5\%$ to $25.0\%$ over reconstructed CF-JEPA. In a separate comparison across six models and two scratch-training seeds, LATER achieves the lowest mean physical readout error in both clean and distracted views, reducing along-push mean absolute error (MAE) under distractors by $20.2\%$ relative to CF-JEPA. Controlled history interventions show that later predictions depend on earlier physical evidence even when query observations and actions are fixed, with larger effects during pushing than waiting.
\end{abstract}

\begin{figure*}[t]
  \centering
  \includegraphics[width=0.7\textwidth]{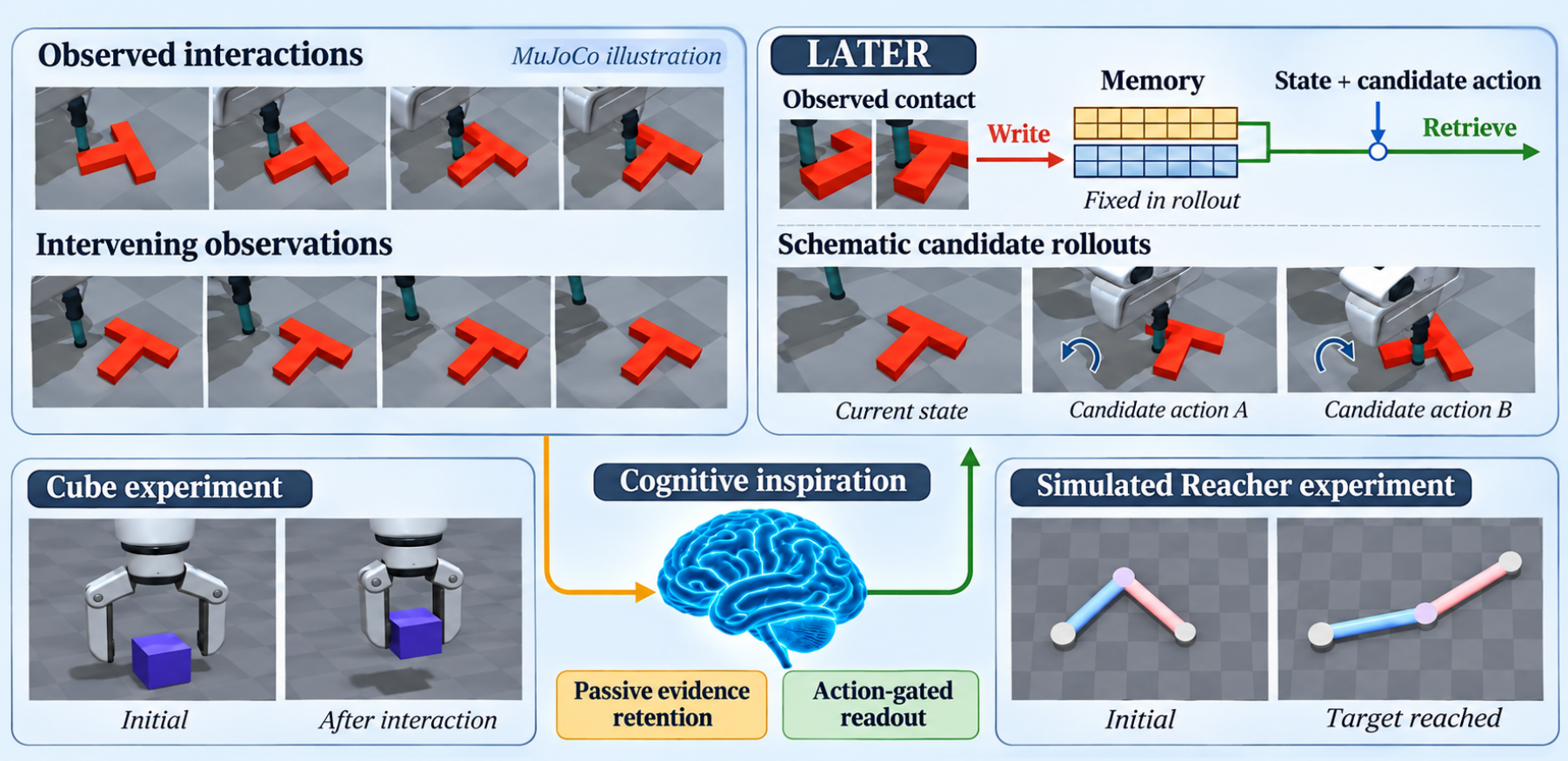}
  \caption{\textbf{LATER in a pushing task.}
  An early push reveals how the T-shaped object responds to contact,
  but later frames may no longer show that response.
  Inspired by working memory and cognitive science, LATER stores
  evidence from observed interactions and reads it using the
  current state and candidate action. Bottom: simulated Cube and
  Reacher examples flank the retention--readout analogy.}
  \label{fig:teaser}
\end{figure*}

\section{Introduction}

Joint-Embedding Predictive Architectures (JEPAs) predict environmental transitions within compact latent spaces~\cite{assran2023ijepa,assran2025vjepa2}. CF-JEPA~\cite{byrd2026cfjepa} improves robustness to visual distractors by decomposing the latent space into a controllable stream ($c$) and an action-adversarial stream ($u$). Isolating controllable dynamics from background noise does not, however, guarantee that earlier physical cues remain available: intervening observations can attenuate information whose predictive value emerges only at a later stage.

Recent JEPA-based world models learn action-conditioned dynamics from finite visual histories~\cite{maes2026lewm,ivashkov2026sensorimotor,byrd2026cfjepa}. This leaves a separate question: can evidence from an earlier interaction still inform a prediction after it is no longer visible in the recent observations? A brief push can reveal surface friction; an exploratory traversal can expose a passage rule. Once motion ceases or the scene resets, the recent images may no longer reveal the property that governs the next action.

Cognitive working memory separates maintenance from selective behavioral readout~\cite{chatham2014outputgating,panichello2021memoryselection}. Information outside the current focus of attention can remain available and be reactivated later~\cite{stokes2015activitysilent,rose2016reactivation}. This distinction motivates separate writing and reading operations. We address two challenges: \textbf{retaining interaction evidence} through intervening observations, and \textbf{reading it for a candidate action} without allowing hypothetical outcomes to overwrite the observations used to evaluate them.

To this end, we propose \textbf{LATER} (\textbf{L}earning \textbf{A}ction-Conditioned \textbf{T}emporal \textbf{E}vidence \textbf{R}etrieval), a predictive framework that integrates persistent memory with action-conditioned retrieval atop a CF-JEPA backbone~\cite{byrd2026cfjepa} (Fig.~\ref{fig:teaser}). LATER introduces a partitioned memory bank wherein interaction memories remain strictly frozen during counterfactual rollouts. Only completed interactions update these banks; hypothetical trajectories cannot overwrite the evidence used to evaluate them.

The \textit{response bridge} uses the joint state--action query $(c_t,a_t)$ to attend to the frozen memory and inject a gated rank-8 correction into the controllable prediction (Fig.~\ref{fig:architecture}(c)). Zero-initialized gates retain baseline behavior at initialization but block primary-loss gradients to the memory writer. Multi-cut auxiliary supervision provides a direct training path while the gates are closed; its head is discarded during planning.

To systematically evaluate delayed evidence retrieval, we construct an automated synthesis pipeline for multi-stage physical environments, creating \textbf{Two-Stage PushT} and \textbf{Directional TwoRoom}. An initial probe supplies evidence before an inter-stage reset and a later query; transitions across the reset are excluded from prediction targets. Our evaluation separates two questions. Public Reacher, PushT, TwoRoom, and Cube tasks test basic control under matched budgets. The synthesized tasks then test whether a later action benefits from evidence outside the recent window. This distinguishes retaining baseline competence from exploiting delayed physical information. On distracted Two-Stage PushT, LATER raises terminal success rate (SR) from 15.5\% to 25.0\% and lowers primary-wait physical readout mean absolute error (MAE) by 27.9\%. Directional TwoRoom first-hit SR rises from 15.0\% to 26.5\%. A separate rendering stress test keeps the two-distractor-trained Reacher checkpoint fixed and obtains 94--100\% SR with 8--64 distractors. Controlled history interventions then isolate the effect of earlier physical evidence on later predictions.

We make three main contributions. First, motivated by the retention--readout distinction in working memory, we study physical evidence whose predictive value depends on a later action. Second, LATER combines partitioned real-transition memories, an action-conditioned response bridge, and auxiliary supervision for zero-gated memory pathways. Third, an automated pipeline synthesizes paired, multi-stage interactions. Matched control comparisons, physical readouts, ablations, and history interventions measure how earlier evidence affects later predictions.

\section{Related Work}

\subsection{JEPA World Models}

JEPAs predict abstract features rather than reconstruct pixels~\cite{assran2025vjepa2,byrd2026cfjepa}. I-JEPA models static images~\cite{assran2023ijepa}; later work extends latent prediction to action-conditioned dynamics. DINO-WM learns dynamics over pretrained patch features for planning~\cite{zhou2025dinowm}. Sparse Imagination reduces the number of visual tokens used during imagined rollouts to lower planning
cost~\cite{chun2026sparseimagination}. LeWM uses latent-distribution regularization to prevent collapse~\cite{balestriero2025lejepa,maes2026lewm}, while H-JEPA introduces multiscale prediction~\cite{zhang2026hjepa}. RC-aux studies the gap between prediction and planning through multi-horizon prediction and budget-conditioned reachability~\cite{li2026rcaux}. Our focus is whether evidence from a past interaction remains available after it leaves the recent context window.

\subsection{Distractor-Resilient Representations and Context Adaptation}

Latent compression alone does not guarantee a separation between action-relevant dynamics and nuisance variation. CF-JEPA uses controllable and action-adversarial branches to improve robustness~\cite{byrd2026cfjepa}. Denoised MDPs factorize
state transitions according to action influence and task
relevance~\cite{wang2022denoisedmdps}. Another line of work addresses
non-stationary environments or hidden physical parameters by
extracting context from historical interaction sequences for rapid
dynamics adaptation, as in CaDM~\cite{lee2020cadm} and
CoDA~\cite{kirchmeyer2022coda}, or by using expected task impact to guide
the compression of historical experience, as in
MemoWM~\cite{zeng2026memowm}. Spatial factorization and recurrent history encoding~\cite{cho2014encoderdecoder} address which variation to retain and how to summarize it. LATER separates real-transition writing from retrieval during candidate rollouts. Generic history models and matched memory--bridge controls test this separation.

\subsection{Working Memory Dissociation in Cognitive Science}

Our design is inspired by the dissociation in cognitive working
memory between maintaining information and selecting it for
behavior~\cite{chatham2014outputgating,panichello2021memoryselection}.
Activity-silent accounts propose that information can be
maintained without sustained content-specific neural activity
and later reactivated~\cite{stokes2015activitysilent,rose2016reactivation}.
Complementary work shows that separating sensory and memory
representations into orthogonal subspaces can reduce
interference~\cite{libby2021rotationalmemory}.
Inspired by this functional separation, LATER decouples passive
evidence retention across transitions from active
action-conditioned readout during rollouts.

\section{Method}
\label{sec:method}

\subsection{Overall Framework}
\label{sec:method_overview}

LATER predicts later action outcomes from recent context and earlier interaction evidence, such as friction revealed before a resting interval.

\begin{figure*}[t]
    \centering
    \includegraphics[width=0.7\textwidth]{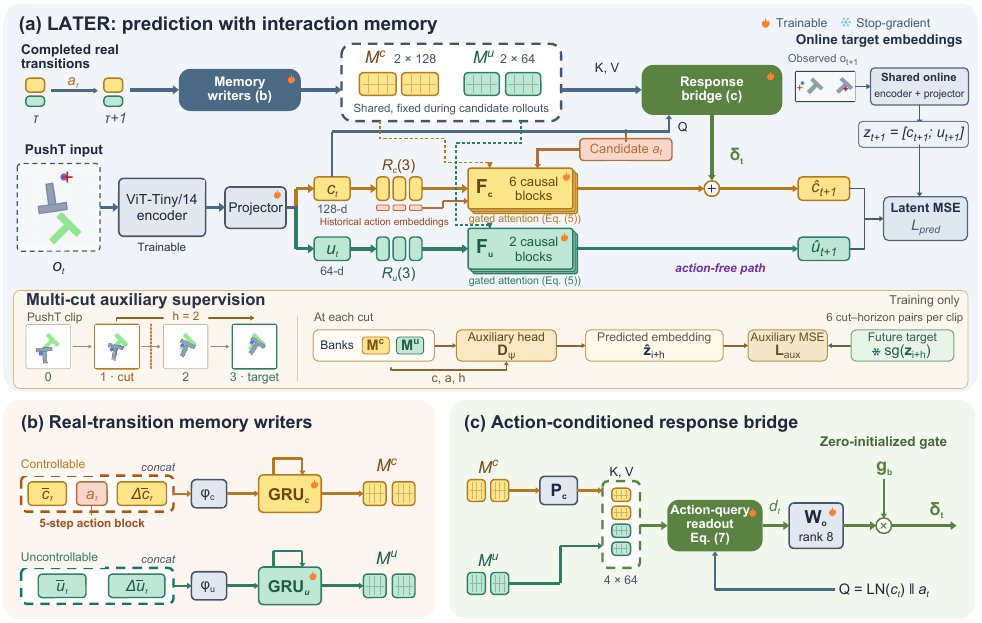}
    \caption{
    \textbf{Architecture of LATER.}
    (a) The trainable CF backbone receives recent tokens and zero-gated memory inputs. The bridge corrects only the controllable prediction; no pixel decoder is used.
    (b) Separate GRU writers summarize completed real transitions. Both banks stay fixed during candidate rollouts.
    (c) A state--action query reads both banks for a gated rank-8 residual (Eqs.~\eqref{eq:bridge_response}--\eqref{eq:bridge_output}). Multi-cut supervision trains the writers; its head is discarded for planning.
    }
    \label{fig:architecture}
\end{figure*}

Let $H_t$ contain completed, valid transitions before time $t$ and $R_t$ denote recent context. Given candidate action $a_t$, LATER forms $M_t=\mathcal{W}(H_t)$ and predicts from $(R_t,a_t,M_t)$. It extends CF-JEPA~\cite{byrd2026cfjepa} with two recurrent writers, a response bridge reading both banks, and auxiliary supervision within each training clip (Fig.~\ref{fig:architecture}). Only completed real transitions update memory; candidate rollouts read it without modification. Paired-scene interventions test the contribution of history while holding the query observation and action fixed.

\subsection{Partitioned Memory from Observed Transitions}
\label{sec:method_memory}

\paragraph{Factorized latent states.}
Following CF-JEPA, a trainable online ViT-Tiny/14 encoder and projector map each observation $o_t$ to
\begin{equation}
z_t=E_\theta(o_t)=[c_t;u_t],\quad
c_t\in\mathbb{R}^{d_c},\quad u_t\in\mathbb{R}^{d_u}.
\label{eq:latent_split}
\end{equation}
Unless stated otherwise, $(d_c,d_u)=(128,64)$. The inverse-dynamics objective encourages action information in $c_t$; gradient reversal discourages it in $u_t$. Each latent transition spans five raw environment actions, concatenated as $a_t\in\mathbb{R}^{d_a}$, where $d_a$ is five times the task's raw action dimension. Semicolons denote concatenation.

\paragraph{Real-transition writers.}
Each update uses an observed transition $(o_t,a_t,o_{t+1})$. With layer normalization (LN), let $\bar c_t=\operatorname{LN}(c_t)$ and $\bar u_t=\operatorname{LN}(u_t)$. We form two writer inputs:
\begin{equation}
\begin{aligned}
e_t^c&=\phi_c([\bar c_t;\,a_t;\,\bar c_{t+1}-\bar c_t]),\\
e_t^u&=\phi_u([\bar u_t;\,\bar u_{t+1}-\bar u_t]),
\end{aligned}
\label{eq:memory_inputs}
\end{equation}
where $\phi_c,\phi_u$ are multilayer perceptrons (MLPs). Two gated recurrent units (GRUs)~\cite{cho2014encoderdecoder} accumulate these inputs:
\begin{equation}
m_{t+1}^s=\operatorname{GRU}_s(e_t^s,m_t^s),
\qquad s\in\{c,u\}.
\label{eq:memory_gru}
\end{equation}
Their hidden states are reshaped into $M_t^c\in\mathbb{R}^{2\times d_c}$ and $M_t^u\in\mathbb{R}^{2\times d_u}$. The $c$ writer records the action and observed controllable change; the $u$ writer receives no explicit action. The partition follows the CF objectives rather than a strict separation of physical properties. The bridge reads both banks, allowing either stream to inform the controllable correction.

\subsection{Action-Conditioned Retrieval and Prediction}
\label{sec:method_retrieval}

\paragraph{Memory-aware prediction.}
Let $R_t^c$ and $R_t^u$ denote recent latent-token windows, each containing three tokens. We retain the CF-JEPA predictor structure: $F_c$ has six causal action-conditioned Transformer blocks, while $F_u$ has two causal blocks without action input. The preliminary predictions are
\begin{equation}
\begin{aligned}
b_{t+1}^c&=F_c([M_t^c;R_t^c],\widetilde A_t;g_c),\\
b_{t+1}^u&=F_u([M_t^u;R_t^u];g_u),
\end{aligned}
\label{eq:preliminary_prediction}
\end{equation}
where $\widetilde A_t$ assigns the executed or candidate action conditions to the recent controllable tokens, and learned conditions to memory tokens.

We introduce a gated residual that adjusts recent-token attention using the difference between attention over recent tokens and attention over all available tokens:
\begin{equation}
\mathcal{A}_s=
\mathcal{A}_s^{\mathrm{recent}}+
g_s\bigl(\mathcal{A}_s^{\mathrm{all}}-\mathcal{A}_s^{\mathrm{recent}}\bigr),
\qquad s\in\{c,u\}.
\label{eq:memory_attention_gate}
\end{equation}
Both attention terms are output representations for the same recent-token queries, with identical dimensions and shared projections; only $\mathcal{A}_s^{\mathrm{all}}$ includes memory keys and values. The scalar gates $g_c,g_u$ start at zero.

\paragraph{Response bridge.}
The response bridge adds a direct state--action query over both banks. A learned $P_c\in\mathbb{R}^{d_u\times d_c}$ projects each $c$-memory token, giving
\begin{equation}
B_t=[P_c(M_t^c);M_t^u]\in\mathbb{R}^{4\times d_u},
\qquad v_t=[\operatorname{LN}(c_t);a_t].
\label{eq:bridge_query}
\end{equation}
The bridge retrieves a memory summary by cross-attention and combines it with the query through an eight-dimensional bottleneck:
\begin{equation}
\begin{aligned}
r_t&=\operatorname{Attention}(W_Qv_t,B_t,B_t),\\
d_t&=\tanh\!\left(W_m\operatorname{LN}(r_t)\right)
          \odot (W_vv_t),\qquad d_t\in\mathbb{R}^{8}.
\end{aligned}
\label{eq:bridge_response}
\end{equation}
Attention includes learned key and value projections. The maps $W_Q$, $W_m$, and $W_v$ have shapes $d_u\times(d_c+d_a)$, $8\times d_u$, and $8\times(d_c+d_a)$, respectively. Only the controllable prediction receives the correction:
\begin{equation}
\hat c_{t+1}=b_{t+1}^c+g_bW_od_t,\qquad
\hat u_{t+1}=b_{t+1}^u.
\label{eq:bridge_output}
\end{equation}
The learned output projection $W_o\in\mathbb{R}^{d_c\times8}$ places the correction in an at-most-eight-dimensional subspace. The scalar gate $g_b$ starts at zero. With all gates zero and matched shared weights and token positions, LATER recovers the reconstructed recent-only CF-JEPA predictor in deterministic evaluation.

During a candidate rollout, $(c_t,a_t)$ changes with the predicted trajectory, but both banks stay fixed. Different actions can read the same observed evidence without writing hypothetical consequences back into it.

\textbf{Value of action-conditioned memory.}
Consider the scalar surrogate $X=\rho Q+N$ and
$Y=I\gamma AQ+\xi$, where $Q,A\in\{-1,1\}$ are
balanced physical-factor and action variables, and
$I\sim\mathrm{Bernoulli}(p)$ indicates activation.
Here $\rho$ is retained evidence strength and $\gamma$ is the activated response coefficient. Assume mutually independent sources and centered
finite-variance noises $N,\xi$, with $\operatorname{Var}(N)=\nu$.
For the linear readout $\hat Y=bIAX$, define
$\mathcal{R}(b)=\frac12\mathbb{E}[(Y-bIAX)^2]$.
For $p>0$ and $\rho^2+\nu>0$, completing the square gives
$b^\star=\gamma\rho/(\rho^2+\nu)$ and
\begin{equation*}
\begin{aligned}
\mathcal{R}(b)-\mathcal{R}(0)
&=\frac{p}{2}\bigl[(\rho^2+\nu)b^2-2\gamma\rho b\bigr],\\
G_{\mathrm{lin}}
&=\frac{p\gamma^2}{2}\frac{\rho^2}{\rho^2+\nu}
 =\frac{p\gamma^2}{2}\operatorname{Corr}^2(X,Q),
\end{aligned}
\end{equation*}
where $G_{\mathrm{lin}}=\mathcal{R}(0)-\mathcal{R}(b^\star)$.
Inactive intervals ($I=0$) have no memory-dependent response; activated queries benefit from retained evidence. For fixed $p,\gamma$, the gain increases with $\operatorname{Corr}^2(X,Q)$.

\subsection{Training Objectives}
\label{sec:method_training}

\paragraph{Base predictive objective.}
We retain the training losses of our CF-JEPA reconstruction:
\begin{equation}
\mathcal{L}_{\mathrm{CF}}=
\mathcal{L}_{\mathrm{pred}}
+\alpha\mathcal{L}_{\mathrm{inv}}
+\mathcal{L}_{\mathrm{adv}}^{\mathrm{GRL}}
+0.09\mathcal{L}_{\mathrm{SIGReg}}.
\label{eq:cf_loss}
\end{equation}
The prediction term is mean squared error (MSE) between predicted and online target features. Inverse dynamics predicts actions from consecutive $c$ states; the gradient reversal layer (GRL) adversary uses consecutive $u$ states. Sketched Isotropic Gaussian Regularization (SIGReg)~\cite{balestriero2025lejepa} constrains projected features. The inverse weight $\alpha$ is task-specific ($\alpha=3$ for PushT). Targets use the online encoder, without an exponential moving average.

\paragraph{Multi-cut auxiliary supervision.}
At the beginning of training, the zero-initialized gates prevent the main prediction loss from updating the memory writers. To train this pathway directly, we predict observed future embeddings from several cuts of each four-frame clip $(o_0,o_1,o_2,o_3)$. For cut $i\in\{0,1,2\}$ and horizon $1\leq h\leq 3-i$, an auxiliary head $D_\psi$ uses the available memories, current controllable state, logged action suffix, and horizon index:
\begin{equation}
D_{i,h}=D_\psi(M_i^c,M_i^u,c_i,a_{i:i+h-1},h).
\label{eq:aux_predictor}
\end{equation}
Each $M_i$ contains only real transitions completed by cut $i$; future states appear only as detached targets. The three cuts yield six cut--horizon pairs. We optimize
\begin{equation}
\begin{aligned}
\mathcal{L}_{\mathrm{aux}}
&=\frac{1}{6}\sum_{i=0}^{2}\sum_{h=1}^{3-i}
\operatorname{MSE}\!\left(D_{i,h},
\operatorname{sg}(z_{i+h})\right),\\
\mathcal{L}_{\mathrm{total}}
&=\mathcal{L}_{\mathrm{CF}}+0.1\mathcal{L}_{\mathrm{aux}},
\end{aligned}
\label{eq:training_objective}
\end{equation}
where $\operatorname{sg}$ stops gradients through the targets only. The auxiliary head is removed for planning; the writers, memory banks, and response bridge remain.

\paragraph{Cold-start gradient path.}
Let $\theta_M$ contain the writer MLPs and GRUs. At $g_c=g_u=g_b=0$, their primary-loss gradient is zero, but the auxiliary head still reads written memories. Auxiliary supervision can therefore update the writers before the gates open, provided its targets depend on earlier transitions.

\section{Experiments}
\label{sec:experiments}

\begin{table*}[!t]
\centering
\caption{\textbf{Control success rates (\%).}
Ordinary PushT: E7, 150 cases, clean-trained weights shared
across views; others: E10, per-view training, 50 ordinary
or 200 hard cases.
LeWM/DINO-WM lead on clean ordinary/hard PushT, respectively;
LATER leads both under two distractors.
First-hit/terminal score any/final goal hit.
Single-Cube resets are unpaired; the rule separates external
baselines from our variants.}
\label{tab:ordinary_push}
\label{tab:control_results}
\footnotesize
\setlength{\tabcolsep}{2.5pt}
\renewcommand{\arraystretch}{0.97}
\begin{minipage}[t]{0.49\textwidth}
\vspace{0pt}
\begin{tabular*}{0.98\linewidth}{@{\extracolsep{\fill}}lrrrr@{}}
\toprule
\multicolumn{5}{@{}l}{\textit{(a) PushT: ordinary first-hit / hard terminal SR}}\\
& \multicolumn{2}{c}{Ordinary, E7} & \multicolumn{2}{c}{Hard, E10}\\
\cmidrule(lr){2-3}\cmidrule(l){4-5}
Model & $N_0$ & $N_2$ & $N_0$ & $N_2$\\
\midrule
CF-JEPA~\cite{byrd2026cfjepa} & 45.3 & 5.3 & 7.5 & 15.5\\
LeWM~\cite{maes2026lewm} & \textbf{74.7} & 8.0 & 0.0 & 2.0\\
SMWM~\cite{ivashkov2026sensorimotor} & 44.7 & 10.7 & 0.0 & 0.0\\
DINO-WM~\cite{zhou2025dinowm} & 19.3 & 6.0 & \textbf{43.0} & 12.0\\
Sparse Imagination (adapt.)~\cite{chun2026sparseimagination}
& 18.0 & 6.7 & 17.5 & 10.0 \\
RC-aux~\cite{li2026rcaux} & 8.7 & 5.3 & 39.0 & 2.0\\
\midrule
Ungated C11 & 50.0 & 8.0 & 14.5 & 15.5\\
Zero-init R & 42.0 & 10.0 & 4.0 & 22.5\\
\textbf{LATER} & 58.7 & \textbf{12.7} & 10.5 & \textbf{25.0}\\
\bottomrule
\end{tabular*}

\end{minipage}\hfill
\begin{minipage}[t]{0.49\textwidth}
\vspace{0pt}

\begin{tabular*}{0.98\linewidth}{@{\extracolsep{\fill}}lrr@{}}
\toprule
\multicolumn{3}{@{}l}{\textit{(b) Reacher, E10: first-hit SR}}\\
Model & $N_0$ & $N_2$\\
\midrule
CF-JEPA~\cite{byrd2026cfjepa} & \textbf{100.0} & 100.0\\
LeWM~\cite{maes2026lewm} & \textbf{100.0} & 16.0\\
SMWM~\cite{ivashkov2026sensorimotor} & \textbf{100.0} & 4.0\\
H-JEPA-2L~\cite{zhang2026hjepa} & 98.0 & 2.0\\
CaDM-visual~\cite{lee2020cadm} & \textbf{100.0} & 0.0\\
RC-aux~\cite{li2026rcaux} & 22.0 & 8.0\\
\textbf{LATER} & 98.0 & \textbf{100.0}\\
\end{tabular*}

\smallskip
\begin{tabular*}{0.98\linewidth}{@{\extracolsep{\fill}}lrrrr@{}}
\toprule
\multicolumn{5}{@{}l}{\textit{(c) TwoRoom and Cube, E10}}\\
& \multicolumn{2}{c}{CF-JEPA~\cite{byrd2026cfjepa}} & \multicolumn{2}{c}{\textbf{LATER}}\\
Task / criterion & $N_0$ & $N_2$ & $N_0$ & $N_2$\\
\midrule
TwoRoom / first-hit & 86.0 & 96.0 & \textbf{86.0} & \textbf{98.0}\\
Cube / first-hit & 54.0 & \textbf{60.0} & \textbf{54.0} & 56.0\\
Directional / first-hit & 34.0 & 15.0 & \textbf{35.0} & \textbf{26.5}\\
Directional / terminal & 7.5 & 9.0 & \textbf{7.5} & \textbf{14.0}\\
\bottomrule
\end{tabular*}
\end{minipage}
\end{table*}

\begin{figure*}[t]
\centering
\includegraphics[width=0.7\textwidth]{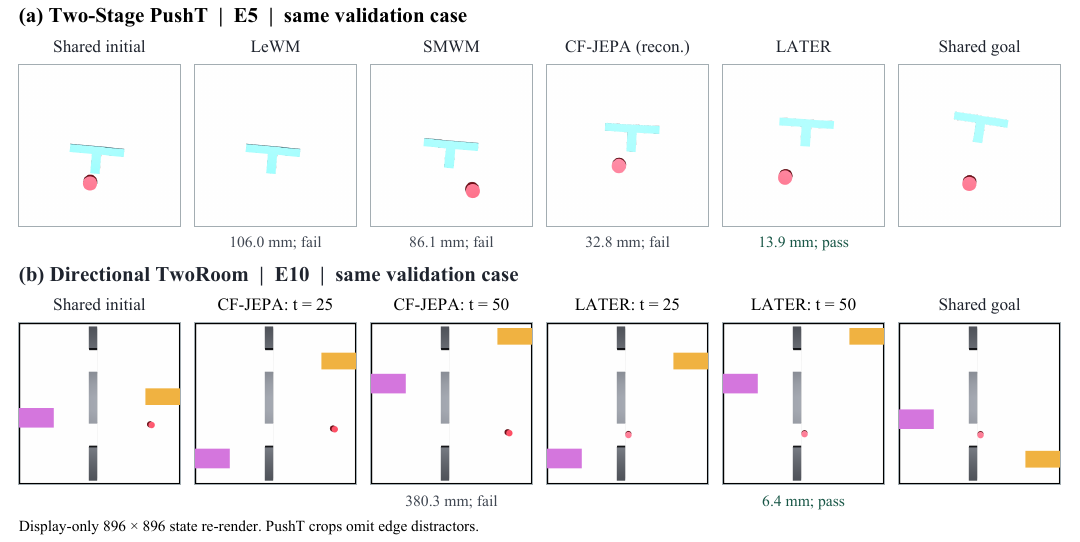}
\caption{\textbf{Matched simulator examples.} PushT E5 (top) and TwoRoom E10 (bottom) share initial state, goal and distractors within each row. These are the first complete cases solved by LATER but not the displayed controls. LeWM/SMWM are matched task-trained adaptations.}
\label{fig:sota_qualitative}
\end{figure*}

\begin{figure}[!t]
  \centering
  \includegraphics[width=0.9\linewidth]{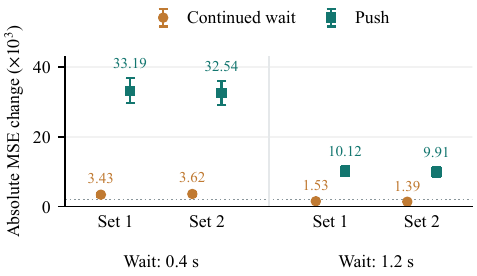}
  \caption{\textbf{Predictions depend on earlier physical evidence.} Absolute prediction-MSE change after a memory-direction swap. Sets 1/2 use distractor-rendering seeds 42/104729 with the same 51 scene pairs and weights. Whiskers are scene-pair-bootstrap 95\% intervals. The dotted margin, 0.002003, is 5\% of development waiting MSE; both waits share one axis.}
  \label{fig:action_dependent_memory}
\end{figure}

\begin{table*}[!t]
\centering
\caption{\textbf{Earlier evidence improves physical prediction.} Frozen E10; 100 confirmation scenes, six pushes each. MAE: along-push error (mm); Acc.: 10-mm endpoint accuracy (\%). (a,c): original cohort; (b): two scratch-training seeds (3072/3073). Readouts use separate fit/development scenes. In (c), brackets are scene-paired 95\% CIs; the last column gives seed means and sample SDs.}
\label{tab:downstream_push}
\label{tab:scratch_readout}
\label{tab:readout_seeds}
\footnotesize
\setlength{\tabcolsep}{2.5pt}
\renewcommand{\arraystretch}{0.97}
\begin{minipage}[t]{0.49\textwidth}
\vspace{0pt}
\centering
\begin{tabular*}{0.98\linewidth}{@{\extracolsep{\fill}}lrrrr@{}}
\toprule
\multicolumn{5}{@{}l}{\textit{(a) Original $N_2$ cohort: native predictions}}\\
Metric & CF-JEPA & LeWM & SMWM & \textbf{LATER}\\
& \cite{byrd2026cfjepa} & \cite{maes2026lewm} & \cite{ivashkov2026sensorimotor} & \\
\midrule
0.4-s MAE $\downarrow$ & 23.70 & 28.66 & 23.59 & \textbf{17.08}\\
0.4-s Acc. $\uparrow$ & 17.3 & 15.0 & 16.5 & \textbf{33.0}\\
1.2-s MAE $\downarrow$ & 23.71 & 28.42 & 23.60 & \textbf{22.33}\\
Held-out amplitude MAE $\downarrow$ & 25.79 & 26.27 & 25.74 & \textbf{17.20}\\
Along-push $R^2$ $\uparrow$ & 0.632 & 0.383 & 0.633 & \textbf{0.783}\\
Along-push Pearson $r$ $\uparrow$ & 0.797 & 0.644 & 0.796 & \textbf{0.885}\\
\midrule
\multicolumn{5}{@{}l}{Primary MAE: 27.9\% lower; 85/100 scenes improve over CF.}\\
\end{tabular*}
\end{minipage}\hfill
\begin{minipage}[t]{0.49\textwidth}
\vspace{0pt}
\centering
\begin{tabular*}{0.98\linewidth}{@{\extracolsep{\fill}}lrrrr@{}}
\toprule
\multicolumn{5}{@{}l}{\textit{(b) Scratch-trained cohort: two-seed means, 0.4-s wait}}\\
& \multicolumn{2}{c}{Clean ($N_0$)} & \multicolumn{2}{c}{Distracted ($N_2$)}\\
\cmidrule(lr){2-3}\cmidrule(l){4-5}
Model & MAE $\downarrow$ & Acc. $\uparrow$ & MAE $\downarrow$ & Acc. $\uparrow$\\
\midrule
CF-JEPA~\cite{byrd2026cfjepa} & 23.68 & 15.9 & 23.68 & 17.1\\
LeWM~\cite{maes2026lewm} & 24.27 & 17.6 & 28.35 & 17.1\\
SMWM~\cite{ivashkov2026sensorimotor} & 23.66 & 16.0 & 23.91 & 16.3\\
H-JEPA-2L~\cite{zhang2026hjepa} & 24.40 & 15.0 & 27.57 & 18.3\\
CaDM-visual~\cite{lee2020cadm} & 24.21 & 17.3 & 28.69 & 14.9\\
\textbf{LATER} & \textbf{21.46} & \textbf{25.3} & \textbf{18.89} & \textbf{27.7}\\
\midrule
CF--LATER MAE reduction (mm) & \multicolumn{2}{c}{2.22} & \multicolumn{2}{c}{4.79}\\
\end{tabular*}
\end{minipage}

\par\smallskip
\begin{tabular*}{0.99\textwidth}{@{\extracolsep{\fill}}lrrrr@{}}
\toprule
\multicolumn{5}{@{}l}{\textit{(c) Original-cohort replication: 0.4-s MAE; model means include sample SD in the last column}}\\
Model / contrast & Seed 3072 & Seed 3073 & Seed 3074 & Across seeds\\
\midrule
CF-JEPA~\cite{byrd2026cfjepa} & 23.70 & 23.74 & 23.72 & $23.72\pm0.02$\\
\textbf{LATER} & \textbf{17.08} & \textbf{20.29} & \textbf{18.96} & $\mathbf{18.78\pm1.61}$\\
Reduction [95\% CI] & 6.62 [5.32, 7.89] & 3.45 [0.69, 6.04] & 4.76 [3.60, 5.84] & 4.94\\
\bottomrule
\end{tabular*}
\end{table*}

\begin{figure*}[t]
\centering
\includegraphics[width=0.7\textwidth]{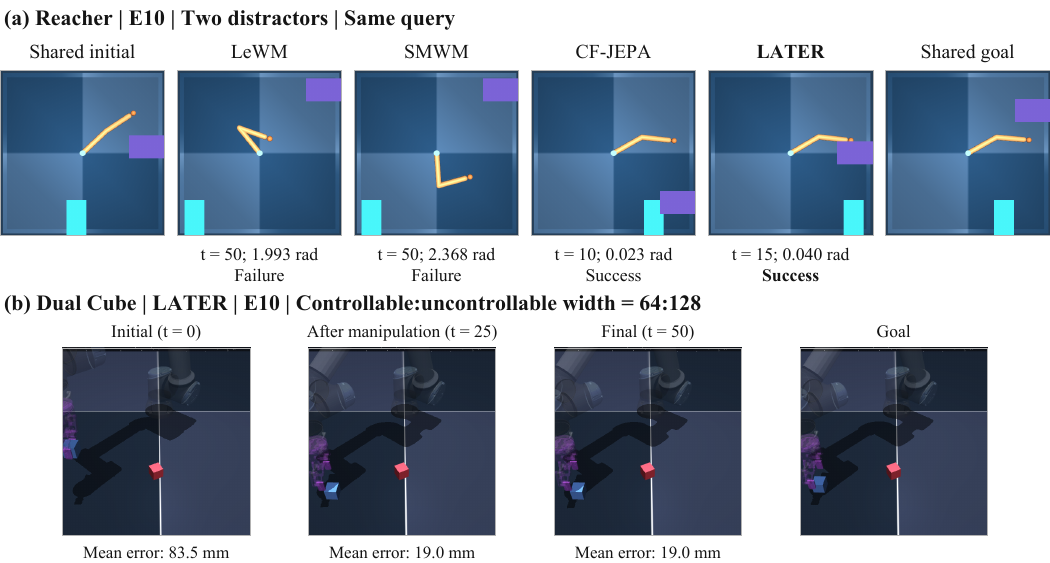}
\caption{\textbf{Reacher and Dual Cube rollouts.} Top: a shared Reacher query at each model's stopping step; maximum joint error must be below 0.05\,rad. Bottom: a successful LATER Dual Cube case ($d_c:d_u=64:128$), labeled by mean position error. Saved states are rendered at 448 pixels; model inputs were 224 pixels. Reacher aggregates appear in Table~\ref{tab:ordinary_push}(b).}
\label{fig:cube_reacher_qualitative}
\end{figure*}

\begin{table}[!htbp]
\centering
\caption{\textbf{Module and memory ablations.} (a) E10 terminal SR (\%): 200 PushT roots; 200 Dual Cube queries from 100 roots (18 already at goal). (b) Fixed-width hard PushT. (c) Original readout, seed 3072: 100 scenes at 2.0\,s; held banks use their 0.4-s states. MAE is in mm.}
\label{tab:main_and_ablation}
\label{tab:memory_ablation}
\footnotesize
\setlength{\tabcolsep}{2.5pt}
\renewcommand{\arraystretch}{0.97}
\begin{tabular*}{0.98\linewidth}{@{\extracolsep{\fill}}lrrrr@{}}
\toprule
\multicolumn{5}{@{}l}{\textit{(a) Memory--bridge factorial across tasks}}\\
& \multicolumn{2}{c}{Two-Stage PushT} & \multicolumn{2}{c}{Dual Cube}\\
Model $(M,B)$ & $N_0$ & $N_2$ & $N_0$ & $N_2$\\
\midrule
Recent-write / No bridge $(0,0)$ & 9.5 & 10.5 & \textbf{19.0} & 13.5\\
Persistent-write / No bridge $(1,0)$ & 7.5 & 16.0 & 16.0 & \textbf{18.0}\\
Recent-write / Bridge $(0,1)$ & 4.5 & 19.0 & 17.5 & 17.0\\
Persistent-write / Bridge $(1,1)$ & \textbf{10.5} & \textbf{25.0} & 15.5 & \textbf{18.0}\\
\end{tabular*}

\smallskip
\begin{tabular*}{\linewidth}{@{\extracolsep{\fill}}lrrr@{}}
\toprule
\multicolumn{4}{@{}l}{\textit{(b) Allocation at fixed total width}}\\
Model & $d_c:d_u$ & $N_0$ SR & $N_2$ SR\\
\midrule
CF-JEPA~\cite{byrd2026cfjepa} & $128:64$ & 7.5 & 15.5\\
CF-JEPA~\cite{byrd2026cfjepa} & $64:128$ & 2.0 & 10.5\\
\textbf{LATER} & $128:64$ & \textbf{10.5} & \textbf{25.0}\\
\textbf{LATER} & $64:128$ & 5.5 & 13.0\\
\bottomrule
\end{tabular*}

\smallskip
\begin{tabular*}{0.98\linewidth}{@{\extracolsep{\fill}}lrrrrr@{}}
\multicolumn{6}{@{}l}{\textit{(c) Bank interventions at 2.0\,s}}\\
Intervention & Native & Hold $M^c$ & Hold $M^u$ & Hold both & Zero both\\
\midrule
MAE $\downarrow$ & 23.94 & 17.16 & 23.90 & \textbf{17.15} & 29.72\\
\bottomrule
\end{tabular*}
\end{table}

\begin{table}[!htbp]
\centering
\caption{\textbf{Controlled history interventions.} (a) Calibrated action choice on 51 held-out pairs at a 0.4-s wait. (b) Original E10 LATER, seed 3072: 100 scenes, six pushes each, 0.4-s wait. Query observations, actions, physical targets, and readout heads are fixed. MAE is in mm; Acc. is 10-mm endpoint accuracy (\%). (c) Gains over CF across the original three seeds. (d) 100 paired clean/distracted renders; both models train on $N_2$.}
\label{tab:mechanism_diagnostics}
\footnotesize
\setlength{\tabcolsep}{2.5pt}
\renewcommand{\arraystretch}{0.97}
\begin{tabular*}{0.98\linewidth}{@{\extracolsep{\fill}}lr@{}}
\toprule
\multicolumn{2}{@{}l}{\textit{(a) Memory-direction swap}}\\
History & Correct / 102\\
\midrule
Factual & 65\\
Swapped & 39\\
\end{tabular*}

\smallskip
\begin{tabular*}{0.98\linewidth}{@{\extracolsep{\fill}}lrr@{}}
\toprule
\multicolumn{3}{@{}l}{\textit{(b) Physical evidence versus visual nuisance}}\\
History condition & MAE $\downarrow$ & Acc. $\uparrow$\\
\midrule
Factual & 17.08 & 33.0\\
Changed distractors only & 16.76 & 33.8\\
Opposite friction & 34.15 & 9.8\\
Early probe removed & 26.37 & 18.2\\
\end{tabular*}
\par\smallskip
\begin{tabular*}{\linewidth}{@{\extracolsep{\fill}}lrr@{}}
\toprule
\multicolumn{3}{@{}l}{\textit{(c) Three-seed readout gains}}\\
Query change & Acc. gain (points) & MAE reduction (mm)\\
\midrule
New distractors & 8.8--16.0 & 1.88--7.32\\
Held-out amplitude & 15.0--17.0 & 5.00--8.59\\
\end{tabular*}
\par\smallskip
\begin{tabular*}{\linewidth}{@{\extracolsep{\fill}}lrr@{}}
\toprule
\multicolumn{3}{@{}l}{\textit{(d) Paired-render feature stability}}\\
Metric & CF-JEPA~\cite{byrd2026cfjepa} & \textbf{LATER}\\
\midrule
Clean/distracted cosine $\uparrow$ & 0.910 & \textbf{0.921}\\
\bottomrule
\end{tabular*}
\end{table}

\begin{table}[!t]
\centering
\caption{\textbf{Reacher with additional distractors.} The same $N_2$-trained E10 LATER checkpoint and 50 questions are used throughout. Only evaluation rendering changes. SR uses the 0.05-rad first-hit criterion; steps count executed actions, capped at 50. Edge rectangles may overlap.}
\label{tab:reacher_stress}
\footnotesize
\setlength{\tabcolsep}{2.5pt}
\renewcommand{\arraystretch}{0.97}
\begin{tabular*}{0.98\linewidth}{@{\extracolsep{\fill}}lrrrr@{}}
\toprule
Distractor count & 8 & 16 & 32 & 64\\
\midrule
SR (\%) $\uparrow$ & 100.0 & 98.0 & 94.0 & 98.0\\
Mean executed steps & 11.9 & 14.4 & 16.9 & 17.1\\
\bottomrule
\end{tabular*}
\end{table}

\subsection{Experimental Setup}
\label{sec:exp_pipeline}

\textbf{Tasks and data synthesis.} Ordinary Reacher, PushT, and TwoRoom use public LeWM data~\cite{maes2026lewm}; single-Cube uses OGBench~\cite{park2025ogbench}. Our multi-stage MuJoCo tasks~\cite{todorov2012mujoco} separate a probe from a later query. In \textit{Two-Stage PushT} (hard PushT), adapted from PushT~\cite{chi2023diffusionpolicy}, a demonstration pushes object A before querying a differently shaped object B on the same surface: friction 0.08--0.65, five shape families, mass 0.30\,kg. In \textit{Directional TwoRoom}, an early probe reveals a passage rule before navigation from the opposite room. Dual Cube extends the probe--query structure to two objects for the factorial ablation.

\textbf{Data splits.} We split physical scenes before rendering or sampling clips; all views and actions from a scene remain in one split. Inter-stage reset transitions are excluded from dynamics targets. Clean ($N_0$) and two-distractor ($N_2$) views share the split. History interventions use 87 development scenes and a separate 51-pair cohort, selected before model scoring by simulator outcomes: both friction worlds must have a successful command and different preferred commands.

\textbf{Baselines and training.} All baseline scores are from our task-trained implementations or adaptations. These include reconstructed CF-JEPA~\cite{byrd2026cfjepa}, LeWM, SMWM, and RC-aux~\cite{maes2026lewm,ivashkov2026sensorimotor,li2026rcaux}; H-JEPA-2L transfers the lower level of a two-level predictor~\cite{zhang2026hjepa}, and CaDM-visual uses one visual model with three-step history~\cite{lee2020cadm}. CF-based runs use four-frame clips, 192-dimensional latents, batch 128, and AdamW~\cite{loshchilov2019adamw} at $5\times10^{-5}$. E$k$ denotes $k$ training epochs. Ordinary PushT uses one clean-trained E7 checkpoint for both views; other control tasks use E10 checkpoints trained separately per view. Ungated C11 and Zero-init R omit auxiliary supervision; R adds zero-initialized gates. Table~\ref{tab:main_and_ablation}(a) instead fixes gates and auxiliary supervision. DINO-WM~\cite{zhou2025dinowm} retains patch features; our Sparse Imagination adaptation~\cite{chun2026sparseimagination} keeps a fixed 98 of 196 DINOv1 patches throughout planning.

\textbf{Planning and success criteria.} The cross-entropy method (CEM) uses 300 candidates, 30 iterations, 30 elites, and five model steps. Each model step groups five raw actions; replanning occurs every five raw steps, up to 50. Success rate (SR) is the fraction of queries reaching the goal: first-hit accepts any hit, while terminal evaluates step 50, including trajectories that hit and later leave. Goal tolerances are maximum joint error $<0.05$\,rad (Reacher), joint pusher/object position error $<20$\,pixels and angle $<\pi/9$ (ordinary PushT), $<16$\,pixels (ordinary TwoRoom), $\leq10$\,mm (Directional TwoRoom), $\leq25$\,mm and $\leq0.15$\,rad (hard PushT), and $\leq40$\,mm per cube. CF-based models score controllable latents, LeWM/SMWM full latents, and RC-aux reachability.

\textbf{Evaluation cohorts and uncertainty.} Control uses 150 ordinary-PushT queries, 50 per other public task, and 200 per hard task; Dual Cube has two queries per root. Hard-$N_0$ DINO-WM and Sparse entries are interim estimates from 21 and 63 completed queries. Physical readout uses disjoint 100/40/100 fit/development/confirmation roots, separate from control and history-selection cohorts; both original and scratch-trained comparisons reuse its confirmation roots. Paired 95\% confidence intervals (CIs) resample physical roots with all derived queries together (scene pairs for history interventions); reported sample SDs measure variation across training seeds.

\textbf{Frozen readout protocol.} Following physical dynamics benchmarks~\cite{bauza2019omnipush,tung2023physionpp}, ridge readouts estimate B's displacement from frozen E10 models. Methods share scenes and regularization grids. Inputs concatenate the current latent, predicted change, and query actions: $c_{128}$ for CF/LATER, $z_{192}$ otherwise. Each scene supplies two friction conditions and command amplitudes 0.55/0.70/0.85 after a 0.70-amplitude demonstration; 0.775 is held out from fitting. MAE measures along-push error; accuracy uses a 10-mm Euclidean endpoint tolerance.

\subsection{Benchmark Control and Delayed-Evidence Planning}
\label{sec:exp_main_results}
\label{sec:ordinary_reacher}

\textbf{Ordinary-task competence.} Reacher remains near saturation (Table~\ref{tab:ordinary_push}): CF and LATER both solve 50/50 distracted cases, versus 8/50 for LeWM and 2/50 for SMWM. LATER matches CF on clean TwoRoom and Cube. On ordinary PushT, it improves over CF by 13.3 clean-view points and leads the reported controls under evaluation-time distractors; LeWM leads on clean views. Figure~\ref{fig:cube_reacher_qualitative} shows Reacher and Dual Cube rollouts.

\textbf{Delayed-evidence control.} On distracted Two-Stage PushT, LATER improves terminal SR by 9.5 points over CF (paired 95\% CI $[3.0,16.0]$). Directional TwoRoom first-hit SR improves by 11.5 points ($[5.5,17.5]$). These tasks require a probe from before the query-stage reset. Figure~\ref{fig:sota_qualitative} shows separate E5 PushT and E10 TwoRoom examples; quantitative CF--LATER comparisons use all 200 queries at E10.

\subsection{Physical Dynamics Readout}
\label{sec:downstream_push}

\textbf{Physical outcome prediction.} Table~\ref{tab:downstream_push}(a) and (c) report the original frozen-checkpoint readout and seed replications. LATER lowers primary-wait MAE from 23.70 to 17.08\,mm (27.9\%) and raises 10-mm endpoint accuracy from 17.3\% to 33.0\%. Errors decrease on 85/100 scenes, with positive scene-paired MAE reductions across all three training seeds. These are physical prediction metrics, distinct from closed-loop SR.

\textbf{Cross-model replication.} Averaging two scratch-training seeds across six methods (Table~\ref{tab:scratch_readout}(b)), LATER has the lowest mean MAE and highest mean endpoint accuracy in both views. Mean MAE reductions over CF are 2.22\,mm in $N_0$ and 4.79\,mm in $N_2$.

\subsection{Ablations and Mechanism Analysis}
\label{sec:exp_ablation}
\label{sec:exp_analysis}

\textbf{Memory and response bridge.} In Table~\ref{tab:main_and_ablation}(a), $M$ enables retention beyond the recent window and $B$ enables the action-conditioned bridge; $(1,1)$ is LATER. All four variants retain GRU writers, gates, and auxiliary supervision; $M=0$ rebuilds memory from recent transitions. Distracted PushT gives 16.0/19.0/25.0\% SR for $(1,0)/(0,1)/(1,1)$. Adding the bridge to persistent memory gains 9.0 points ($[3.0,15.5]$). On distracted Dual Cube, LATER gains 4.5 points over $(0,0)$ ($[1.0,8.0]$), matching $(1,0)$.

\textbf{Capacity allocation.} At fixed $d_c+d_u=192$, allocating more dimensions to the controllable stream improves terminal SR in both views for both models (Table~\ref{tab:main_and_ablation}(b)). LATER retains two tokens per bank, totaling 384 persistent scalars at either allocation.

\label{sec:exp_intervention}
\textbf{Physical evidence versus visual nuisance.} Replacing the early demonstration with an opposite-friction history raises displacement MAE from 17.08 to 34.15\,mm; removing the early probe raises it to 26.37\,mm. Changing only historical distractors gives 16.76\,mm (Table~\ref{tab:mechanism_diagnostics}(b)). Query observations, actions, physical targets, and readout heads stay fixed. The contrast isolates sensitivity to the physical information supplied by the earlier interaction rather than its incidental appearance.

\textbf{Action-dependent history use.} Swapping a development-selected memory direction~\cite{geiger2021causalabstractions} lowers calibrated choice accuracy from 65/102 to 39/102 (Table~\ref{tab:mechanism_diagnostics}(a)). The prediction-MSE change is larger during pushing than waiting at both delays and under both distractor sets (Fig.~\ref{fig:action_dependent_memory}): evidence with little effect on waiting predictions becomes relevant when contact resumes. This is the retention--readout distinction motivating LATER, tested by changing stored evidence while holding the query fixed.

\textbf{Retention across waiting intervals.} Holding the $c$ bank at its earlier state lowers 2.0-s MAE from 23.94 to 17.16\,mm, with reductions of 6.71--7.32\,mm across three seeds (seed 3072 in Table~\ref{tab:memory_ablation}(c)). Holding $M^u$ alone has little effect; holding both banks closely matches holding $M^c$. The native benefit diminishes with waiting, but the earlier bank remains useful to the same reader. This identifies subsequent memory updates as a source of lost predictive information in this setting and motivates event-triggered memory consolidation.

\textbf{Robustness and representations.} Readout gains persist under new distractors and a held-out push amplitude (Table~\ref{tab:mechanism_diagnostics}(c)); paired-render cosine similarity is 0.921 versus CF's 0.910 in (d). On the same distracted hard-PushT trajectories, LATER exceeds CF terminal SR by 2.0--9.5 points across 10/20/25/30/50-mm tolerances. The angle threshold stays at 0.15\,rad; 25\,mm remains primary.

\textbf{Transfer to more distractors.} The fixed $N_2$-trained Reacher checkpoint solves at least 47/50 cases with 8--64 distractors (Table~\ref{tab:reacher_stress}). Only rendering changes; the matched baseline comparison remains at $N_2$ (Table~\ref{tab:ordinary_push}(b)).

\textbf{Matched history and cost.} Our Joint-GRU control projects $[z_t;a_t;z_{t+1}-z_t]$ into a 384-state GRU~\cite{cho2014encoderdecoder}. With matched data, CEM budgets, and ten epochs from one CF checkpoint, the reported continuation run gives 17.0\% terminal SR for LATER, 14.0\% for Joint-GRU, and 15.0\% for recent-only CF-JEPA~\cite{byrd2026cfjepa} on 200 $N_2$ scenes after a 2.4-s wait. Joint-GRU and LATER each retain 384 FP32 scalars (1.5\,KiB). Five CEM calls on a shared H100 give median CF/LATER times of 6.76/9.98\,s and allocator peaks of 122/150\,MiB.

\section{Conclusion}
\label{sec:conclusion}

We introduce \textbf{LATER}, a JEPA-based world model that separates storing interaction evidence from retrieving it for actions. Partitioned memory stays fixed during candidate rollouts; auxiliary losses train the memory pathway. LATER retains basic control competence and improves delayed-evidence control and physical prediction. History interventions connect these gains to earlier physical evidence, making the retention--readout distinction from working memory experimentally testable. Future work can improve success on current and harder tasks over longer interactions.

\section*{LLM Usage Statement}
OpenAI ChatGPT and Codex assisted with literature review, experiment design and analysis, code development and debugging, manuscript drafting and editing, and figure preparation. The authors take full responsibility for the manuscript.

{
  \small
  \bibliographystyle{ieeenat_fullname}
  \bibliography{references}

@inproceedings{assran2023ijepa,
  author = {Assran, Mahmoud and Duval, Quentin and Misra, Ishan and Bojanowski, Piotr and Vincent, Pascal and Rabbat, Michael and LeCun, Yann and Ballas, Nicolas},
  title = {Self-Supervised Learning from Images with a Joint-Embedding Predictive Architecture},
  booktitle = {Proceedings of the IEEE/CVF Conference on Computer Vision and Pattern Recognition},
  pages = {15619--15629},
  year = {2023},
  doi = {10.1109/CVPR52729.2023.01499},
  eprint = {2301.08243},
  archivePrefix = {arXiv},
  primaryClass = {cs.CV},
  url = {https://arxiv.org/abs/2301.08243}
}

@misc{assran2025vjepa2,
  author = {Assran, Mido and Bardes, Adrien and Fan, David and Garrido, Quentin and Howes, Russell and Komeili, Mojtaba and Muckley, Matthew and Rizvi, Ammar and Roberts, Claire and Sinha, Koustuv and Zholus, Artem and Arnaud, Sergio and Gejji, Abha and Martin, Ada and Hogan, Francois Robert and Dugas, Daniel and Bojanowski, Piotr and Khalidov, Vasil and Labatut, Patrick and Massa, Francisco and Szafraniec, Marc and Krishnakumar, Kapil and Li, Yong and Ma, Xiaodong and Chandar, Sarath and Meier, Franziska and LeCun, Yann and Rabbat, Michael and Ballas, Nicolas},
  title = {{V-JEPA 2}: Self-Supervised Video Models Enable Understanding, Prediction and Planning},
  year = {2025},
  eprint = {2506.09985},
  archivePrefix = {arXiv},
  url = {https://arxiv.org/abs/2506.09985},
  howpublished = {\href{https://arxiv.org/abs/2506.09985}{arXiv:2506.09985}}
}

@misc{balestriero2025lejepa,
  author = {Balestriero, Randall and LeCun, Yann},
  title = {{LeJEPA}: Provable and Scalable Self-Supervised Learning Without the Heuristics},
  year = {2025},
  eprint = {2511.08544},
  archivePrefix = {arXiv},
  primaryClass = {cs.LG},
  url = {https://arxiv.org/abs/2511.08544},
  howpublished = {\href{https://arxiv.org/abs/2511.08544}{arXiv:2511.08544}}
}

@inproceedings{bauza2019omnipush,
  author = {Bauza, Maria and Alet, Ferran and Lin, Yen-Chen and Lozano-P{\'e}rez, Tom{\'a}s and Kaelbling, Leslie P. and Isola, Phillip and Rodriguez, Alberto},
  title = {{OmniPush}: Accurate, Diverse, Real-World Dataset of Pushing Dynamics with {RGB-D} Video},
  booktitle = {IEEE/RSJ International Conference on Intelligent Robots and Systems},
  year = {2019},
  doi = {10.1109/IROS40897.2019.8967920},
  url = {https://arxiv.org/abs/1910.00618}
}

@misc{byrd2026cfjepa,
  author = {Byrd, Morgan and Wright, Robert and Ha, Sehoon},
  title = {{CF-JEPA}: Improving Robustness of {JEPA} World Models via Controllability Factorization},
  year = {2026},
  eprint = {2610.00727},
  archivePrefix = {arXiv},
  url = {https://arxiv.org/abs/2610.00727},
  howpublished = {\href{https://arxiv.org/abs/2610.00727}{arXiv:2610.00727}}
}

@article{chatham2014outputgating,
  author = {Chatham, Christopher H. and Frank, Michael J. and Badre, David},
  title = {Corticostriatal output gating during selection from working memory},
  journal = {Neuron},
  volume = {81},
  number = {4},
  pages = {930--942},
  year = {2014},
  doi = {10.1016/j.neuron.2014.01.002},
  url = {https://pmc.ncbi.nlm.nih.gov/articles/PMC3955887/}
}

@inproceedings{chi2023diffusionpolicy,
  author = {Chi, Cheng and Feng, Siyuan and Du, Yilun and Xu, Zhenjia and Cousineau, Eric and Burchfiel, Benjamin and Song, Shuran},
  title = {Diffusion Policy: Visuomotor Policy Learning via Action Diffusion},
  booktitle = {Proceedings of Robotics: Science and Systems},
  year = {2023},
  doi = {10.15607/RSS.2023.XIX.026},
  url = {https://roboticsproceedings.org/rss19/p026.html}
}

@inproceedings{cho2014encoderdecoder,
  author = {Cho, Kyunghyun and van Merri{\"e}nboer, Bart and Gulcehre, Caglar and Bahdanau, Dzmitry and Bougares, Fethi and Schwenk, Holger and Bengio, Yoshua},
  title = {Learning Phrase Representations using {RNN} Encoder--Decoder for Statistical Machine Translation},
  booktitle = {Proceedings of the 2014 Conference on Empirical Methods in Natural Language Processing (EMNLP)},
  pages = {1724--1734},
  year = {2014},
  doi = {10.3115/v1/D14-1179},
  url = {https://aclanthology.org/D14-1179/}
}

@inproceedings{geiger2021causalabstractions,
  author = {Geiger, Atticus and Lu, Hanson and Icard, Thomas and Potts, Christopher},
  title = {Causal Abstractions of Neural Networks},
  booktitle = {Advances in Neural Information Processing Systems},
  volume = {34},
  year = {2021},
  url = {https://papers.neurips.cc/paper_files/paper/2021/hash/4f5c422f4d49a5a807eda27434231040-Abstract.html}
}

@inproceedings{ivashkov2026sensorimotor,
  author = {Ivashkov, Petr and Balestriero, Randall and Sch{\"o}lkopf, Bernhard},
  title = {Sensorimotor World Models: Perception for Action via Inverse Dynamics},
  year = {2026},
  eprint = {2606.20104},
  archivePrefix = {arXiv},
  primaryClass = {cs.LG},
  url = {https://arxiv.org/abs/2606.20104},
  booktitle = {Advances in Neural Information Processing Systems}
}

@inproceedings{kirchmeyer2022coda,
  author = {Kirchmeyer, Matthieu and Yin, Yuan and Dona, Jeremie and Baskiotis, Nicolas and Rakotomamonjy, Alain and Gallinari, Patrick},
  title = {Generalizing to New Physical Systems via Context-Informed Dynamics Model},
  booktitle = {Proceedings of the 39th International Conference on Machine Learning},
  series = {Proceedings of Machine Learning Research},
  volume = {162},
  pages = {11283--11301},
  year = {2022},
  url = {https://proceedings.mlr.press/v162/kirchmeyer22a.html}
}

@inproceedings{lee2020cadm,
  author = {Lee, Kimin and Seo, Younggyo and Lee, Seunghyun and Lee, Honglak and Shin, Jinwoo},
  title = {Context-aware Dynamics Model for Generalization in Model-Based Reinforcement Learning},
  booktitle = {Proceedings of the 37th International Conference on Machine Learning},
  series = {Proceedings of Machine Learning Research},
  volume = {119},
  pages = {5757--5766},
  year = {2020},
  url = {https://proceedings.mlr.press/v119/lee20g.html}
}

@inproceedings{li2026rcaux,
  author = {Li, Wenyuan and Li, Guang and Maeda, Keisuke and Ogawa, Takahiro and Haseyama, Miki},
  title = {Predictive but Not Plannable: {RC-aux} for Latent World Models},
  booktitle = {Advances in Neural Information Processing Systems},
  year = {2026},
  eprint = {2605.07278},
  archivePrefix = {arXiv},
  primaryClass = {cs.LG},
  url = {https://arxiv.org/abs/2605.07278v1},
}

@article{libby2021rotationalmemory,
  author = {Libby, Alexandra and Buschman, Timothy J.},
  title = {Rotational dynamics reduce interference between sensory and memory representations},
  journal = {Nature Neuroscience},
  volume = {24},
  number = {5},
  pages = {715--726},
  year = {2021},
  doi = {10.1038/s41593-021-00821-9},
  url = {https://pmc.ncbi.nlm.nih.gov/articles/PMC8102338/}
}

@inproceedings{loshchilov2019adamw,
  author = {Loshchilov, Ilya and Hutter, Frank},
  title = {Decoupled Weight Decay Regularization},
  booktitle = {International Conference on Learning Representations},
  year = {2019},
  url = {https://arxiv.org/abs/1711.05101}
}

@inproceedings{maes2026lewm,
  author = {Maes, Lucas and Le Lidec, Quentin and Scieur, Damien and LeCun, Yann and Balestriero, Randall},
  title = {{LeWorldModel}: Stable End-to-End Joint-Embedding Predictive Architecture from Pixels},
  booktitle = {Advances in Neural Information Processing Systems},
  year = {2026},
  eprint = {2603.19312},
  archivePrefix = {arXiv},
  primaryClass = {cs.LG},
  url = {https://arxiv.org/abs/2603.19312},
}

@article{panichello2021memoryselection,
  author = {Panichello, Matthew F. and Buschman, Timothy J.},
  title = {Shared mechanisms underlie the control of working memory and attention},
  journal = {Nature},
  volume = {592},
  number = {7855},
  pages = {601--605},
  year = {2021},
  doi = {10.1038/s41586-021-03390-w},
  url = {https://pubmed.ncbi.nlm.nih.gov/33790467/}
}

@inproceedings{park2025ogbench,
  author = {Park, Seohong and Frans, Kevin and Eysenbach, Benjamin and Levine, Sergey},
  title = {{OGBench}: Benchmarking Offline Goal-Conditioned {RL}},
  booktitle = {International Conference on Learning Representations},
  year = {2025},
  eprint = {2410.20092},
  archivePrefix = {arXiv},
  primaryClass = {cs.LG},
  url = {https://openreview.net/forum?id=M992mjgKzI}
}

@article{rose2016reactivation,
  author = {Rose, Nathan S. and LaRocque, Joshua J. and Riggall, Adam C. and Gosseries, Olivia and Starrett, Michael J. and Meyering, Emma E. and Postle, Bradley R.},
  title = {Reactivation of latent working memories with transcranial magnetic stimulation},
  journal = {Science},
  volume = {354},
  number = {6316},
  pages = {1136--1139},
  year = {2016},
  doi = {10.1126/science.aah7011},
  url = {https://doi.org/10.1126/science.aah7011}
}

@article{stokes2015activitysilent,
  author = {Stokes, Mark G.},
  title = {{`Activity-silent'} working memory in prefrontal cortex: a dynamic coding framework},
  journal = {Trends in Cognitive Sciences},
  volume = {19},
  number = {7},
  pages = {394--405},
  year = {2015},
  doi = {10.1016/j.tics.2015.05.004},
  url = {https://doi.org/10.1016/j.tics.2015.05.004}
}

@inproceedings{todorov2012mujoco,
  author = {Todorov, Emanuel and Erez, Tom and Tassa, Yuval},
  title = {{MuJoCo}: A physics engine for model-based control},
  booktitle = {2012 IEEE/RSJ International Conference on Intelligent Robots and Systems},
  pages = {5026--5033},
  organization = {IEEE},
  year = {2012},
  doi = {10.1109/IROS.2012.6386109}
}

@inproceedings{tung2023physionpp,
  author = {Tung, Hsiao-Yu and Ding, Mingyu and Chen, Zhenfang and Bear, Daniel and Gan, Chuang and Tenenbaum, Joshua B. and Yamins, Daniel L. K. and Fan, Judith E. and Smith, Kevin A.},
  title = {{Physion++}: Evaluating Physical Scene Understanding that Requires Online Inference of Different Physical Properties},
  booktitle = {Advances in Neural Information Processing Systems},
  volume = {36},
  year = {2023},
  doi = {10.52202/075280-2929},
  eprint = {2306.15668},
  archivePrefix = {arXiv},
  primaryClass = {cs.CV},
  url = {https://arxiv.org/abs/2306.15668}
}

@inproceedings{wang2022denoisedmdps,
  author = {Wang, Tongzhou and Du, Simon S. and Torralba, Antonio and Isola, Phillip and Zhang, Amy and Tian, Yuandong},
  title = {Denoised {MDP}s: Learning World Models Better Than the World Itself},
  booktitle = {Proceedings of the 39th International Conference on Machine Learning},
  series = {Proceedings of Machine Learning Research},
  volume = {162},
  pages = {22591--22612},
  year = {2022},
  url = {https://proceedings.mlr.press/v162/wang22c.html}
}

@misc{zeng2026memowm,
  author = {Zeng, Bingfan and Chen, Zhisheng and Sang, Chenbo and Xie, Zhengwei and Wang, Jinpeng and Guan, Xiangchen and Qian, Rui and Lu, Zheng and Song, Jingwei},
  title = {{MemoWM}: How World Models Change What Agents Need to Remember},
  year = {2026},
  eprint = {2610.10778},
  archivePrefix = {arXiv},
  primaryClass = {cs.LG},
  url = {https://arxiv.org/abs/2610.10778v1},
  howpublished = {\href{https://arxiv.org/abs/2610.10778}{arXiv:2610.10778}}
}

@misc{zhang2026hjepa,
  author = {Zhang, Wancong and Terver, Basile and Rabbat, Michael and LeCun, Yann and Balestriero, Randall},
  title = {{H-JEPA}: End-to-End Learning of Hierarchical World Models for Visual Planning},
  year = {2026},
  eprint = {2610.06805},
  archivePrefix = {arXiv},
  url = {https://arxiv.org/abs/2610.06805},
  howpublished = {\href{https://arxiv.org/abs/2610.06805}{arXiv:2610.06805}}
}

@inproceedings{zhou2025dinowm,
  author = {Zhou, Gaoyue and Pan, Hengkai and LeCun, Yann and Pinto, Lerrel},
  title = {{DINO-WM}: World Models on Pre-trained Visual Features Enable Zero-shot Planning},
  booktitle = {Proceedings of the 42nd International Conference on Machine Learning},
  series = {Proceedings of Machine Learning Research},
  volume = {267},
  pages = {79115--79135},
  year = {2025},
  publisher = {PMLR},
  url = {https://proceedings.mlr.press/v267/zhou25t.html}
}

@inproceedings{chun2026sparseimagination,
  author = {Chun, Junha and Jeong, Youngjoon and Kim, Taesup},
  title = {Sparse Imagination for Efficient Visual World Model Planning},
  booktitle = {International Conference on Learning Representations},
  year = {2026},
  url = {https://arxiv.org/abs/2506.01392}
}
}

\end{document}